\documentclass{article}

\usepackage[final]{neurips_2020}

\usepackage[pdftex]{graphicx}
\usepackage[utf8]{inputenc} % allow utf-8 input
\usepackage[T1]{fontenc}    % use 8-bit T1 fonts
\usepackage{hyperref}       % hyperlinks
\usepackage{url}            % simple URL typesetting
\usepackage{booktabs}       % professional-quality tables
\usepackage{amsfonts}       % blackboard math symbols
\usepackage{nicefrac}       % compact symbols for 1/2, etc.
\usepackage{microtype}      % microtypography

\usepackage{multirow}
\usepackage{bbm}
\usepackage{wrapfig}

\usepackage{color}

\title{Factorizable Graph Convolutional Networks}

\author{%
  Yiding Yang \\ 
  Stevens Institute of Technology \\
  \texttt{yyang99@stevens.edu} \\
    \And
    Zunlei  Feng\\
    Zhejiang University \\
    \texttt{zunleifeng@zju.edu.cn} \\
    \And
    Mingli Song\\
    Zhejiang University \\
    \texttt{brooksong@zju.edu.cn} \\
    \And
    Xinchao Wang \thanks{Corresponding author.}\\
    Stevens Institute of Technology\\
    \texttt{xinchao.wang@stevens.edu}
}

\begin{document}

\maketitle

\begin{abstract}
%Real-world graphs typically 
%encode multiple heterogeneous relations between adjacent nodes,
%such as the different reasons that some products are bought together or 
%the different relations between users in a social network. 
%Graphs have been widely adopted to denote 
Graphs have been widely adopted to 
denote structural connections between entities.
The relations are in many cases heterogeneous, 
but entangled together and denoted merely 
as a single edge  between a pair of nodes.
For example, in a social network graph, 
users in different latent relationships like friends and colleagues,
are usually connected via a bare edge that conceals 
such intrinsic connections.
In this paper, we introduce a
novel graph convolutional network~(GCN),
termed as \emph{factorizable graph convolutional network}~(FactorGCN),
that explicitly
disentangles such intertwined relations 
encoded in a graph.
FactorGCN takes a {simple graph} as input,
and disentangles it into several 
factorized graphs, each of which
represents a latent and disentangled relation among nodes.
The features of the nodes are then aggregated separately in each factorized latent space
to produce disentangled features,
which further leads to better performances for downstream tasks.
We evaluate the proposed FactorGCN both qualitatively and quantitatively
on the synthetic and real-world datasets,
and demonstrate that it yields truly encouraging results
in terms of both disentangling and feature aggregation. 
Code is publicly available at
\url{https://github.com/ihollywhy/FactorGCN.PyTorch}.

%Different from the previous works that all focus on the disentangling the neighbors of the center node, we focus on the disentanglement in the graph level. Specifically, the input graph will be disentangled into several factorized graphs, where each one represents a latent relation among entities.  The features of node will then be aggregated separately in each of the latent relation space, leading to a disentangled features and a better performance of downstream tasks.

%To achieve the graph-level disentanglement, we propose factorized graph convolutional networks that contain several factorized disentangle layers.

%We also define a new metric to evaluate the quantitative performance of graph-level disentanglement. 

\end{abstract}

\section{Introduction}
%In recent years, disentangling has been  widely explored in deep learning. Its goal is to disentangle an entity like a feature vector into several interpretable representations~\citep{bengio2013representation},
%so that XXX.

Disentangling aims to factorize an entity, like a feature vector, into several interpretable components, so that
the behavior of a learning model can be better understood. 
In recent years, many approaches have been proposed towards 
tackling disentangling in deep neural networks 
and have achieved promising results. 
Most prior efforts, however,
have been focused on the disentanglement of 
convolutional neural network~(CNN)
especially the auto-encoder architecture,
where disentangling takes place
during the stage of latent feature generation.
{For example,  VAE~\citep{kingma2013autovae}
restrains the distribution of the latent features
to Gaussian and generates disentangled
representation; $\beta$-VAE~\citep{higgins2017betaVAE} further improves the disentangling by introducing $\beta$ to balance the 
independence constraints and reconstruction accuracy.}

%Using a $\beta>1$  will enforce the auto-encoder to 
%learn a more efficient latent representation and 
%disentangle in an unsupervised manner.

Despite the many prior efforts in CNN disentangling,
there are few endeavors 
toward disentangling in the irregular structural domain, 
where graph convolutional network~(GCN) models are applied. 
Meanwhile, the inherent differences between grid-like data and structural data precludes applying CNN-based disentangling methods to GCN ones.
The works of~\citep{ma2019disentangled,liu2019independence}, 
as pioneering attempts, 
focus on the node-level 
neighbour partition
and ignore the latent multi-relations among nodes.

\begin{figure}
  \centering
  \includegraphics[width=0.95\linewidth]{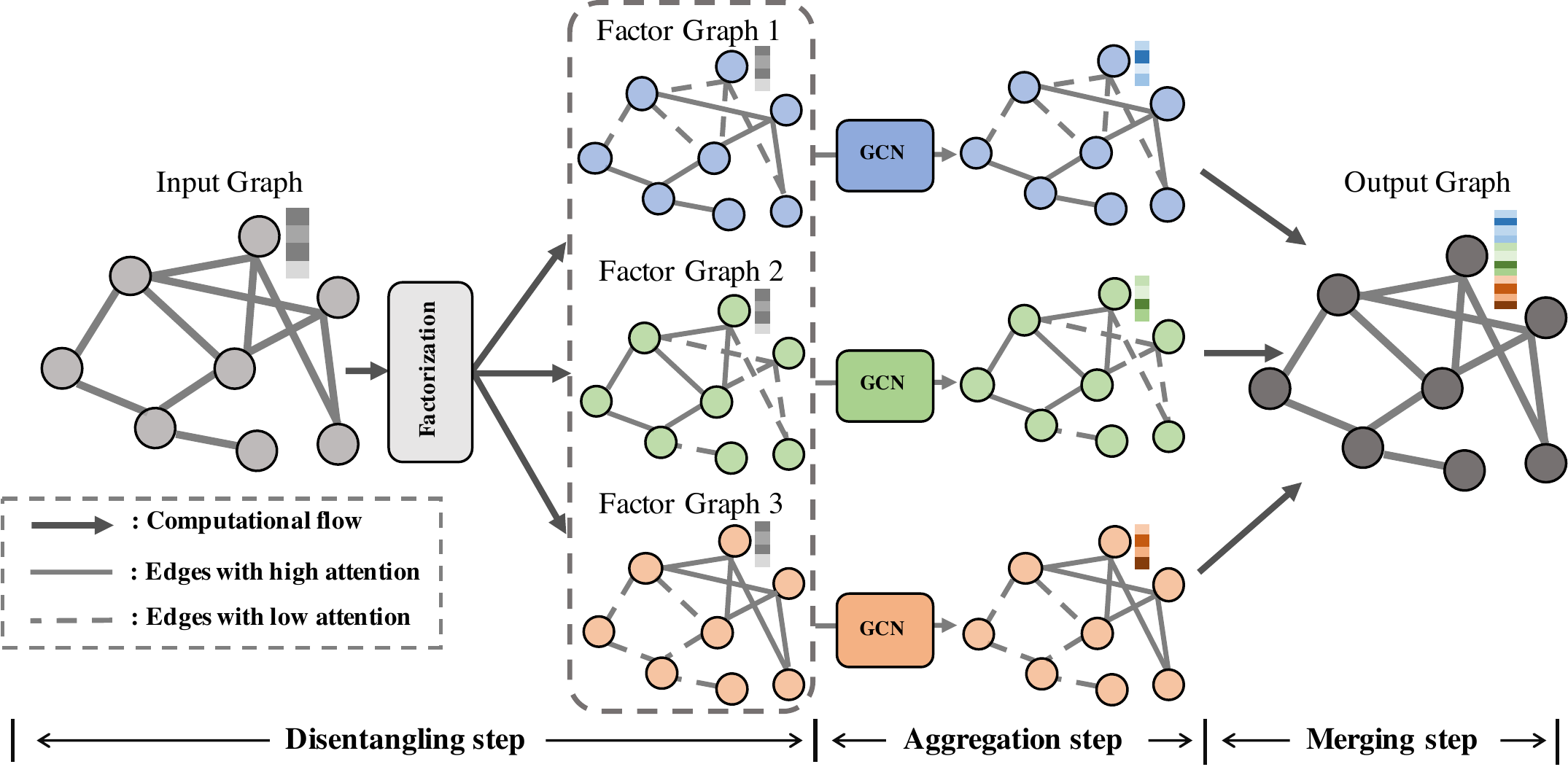}
%   \vspace{-0.5em}
  \caption{Illustration of one layer in the proposed FactorGCN.
    It contains three steps: 
    \emph{Disentangling}, \emph{Aggregation}, and \emph{Merging}.
    In the disentangling step, the input graph is decomposed into several factor graphs, each of which represents a latent relation among nodes. In the aggregation step,  GCNs are applied separately to the derived factor graphs
    and produce the latent features. 
    In the merging step, features from all latent graphs
    are concatenated to form the final features,
    which are block-wise interpretable.}
    \label{fig:factorlayer}
    % \vspace{-1.5em}
\end{figure}

We introduce in this paper a novel GCN, that aims to explicitly conduct graph-level disentangling, based on which convolutional features are aggregated. Our approach, termed as \emph{factorizable graph convolutional network}~(FactorGCN), 
takes as input a \emph{simple graph}, 
and decomposes it into several 
factor graphs, each of which corresponds  
to a disentangled and interpretable relation space, 
as shown in Fig.~\ref{fig:factorlayer}.
Each such graph then undergoes a GCN, tailored to
aggregate features only from one disentangled latent space,
followed by a merging operation that concatenates all derived 
features from disentangled spaces,
so as to produce the final block-wise interpretable features. 
These steps constitute one layer of the proposed FactorGCN.
As the output graph with updated features share the identical
topology as input, nothing prevents us 
from stacking a number of
layers to disentangle the input data at different levels,
yielding a hierarchical disentanglement 
with various numbers of factor 
graph at different levels.
%\xwc{XXX, mention the goal of stacking multiple layer}.

FactorGCN, therefore, 
potentially finds application in a 
wide spectrum of scenarios. 
In many real-world graphs, 
multiple heterogeneous relations 
between nodes are mixed and 
collapsed to one single edge.
In the case of social networks,
two people may be \emph{friends}, \emph{colleagues},
and \emph{living in the same city} simultaneously,
but linked via one single edge that omits such
interconnections;
in the co-purchasing scenario~\citep{co-purchase},
products are bought together for
different reasons like \emph{promotion}, 
and \emph{functional complementary},
but are often ignored in the graph construction.
FactorGCN would, in these cases,
deliver a disentangled and interpretable 
solution towards explaining the underlying 
rationale, and provide discriminant learned features
for the target task.

Specifically, the contributions of FactorGCN are summarized as follows.

\begin{itemize}
  \item {\bf Graph-level Disentangling}. 
  FactorGCN conducts disentangling and produces block-wise
  interpretable node features by analyzing the whole graph
  all at once, during which process the global-level topological semantics,
  such as the higher-order relations between edges and nodes,
  is explicitly accounted for. The disentangled factor graphs
  reveal latent-relation specific interconnections between
  the entities of interests, and yield interpretable features 
  that benefit the downstream tasks. 
  This scheme therefore contrasts to the prior approaches of~\citep{ma2019disentangled,liu2019independence},
  where the {disentanglement takes place only within a local neighborhood, without accounting for global contexts}.
  
  \item {\bf Multi-relation Disentangling}. 
  Unlike prior methods that decode only
  a single attribute for a neighboring node,
  FactorGCN enables multi-relation 
  disentangling, meaning that
  {the center node may aggregate information
  from a neighbour under multiple types of relations}.
  This mechanism is crucial since real-world  
  data may contain various relations among the 
  same pair of entities.
  In the case of a social network graph, for example,
  FactorGCN would produce disentangled results
  allowing for two users to be both \emph{friends}
  and \emph{living in the same city}; such
  multi-relation disentangling is not supported by prior
  GCN methods.
  
  \item {\bf Quantitative Evaluation Metric}.
  Existing quantitative evaluation methods~\citep{eastwood2018framework,burgess2018understanding}
  in the grid domain rely on 
  generative models, like auto-encoder~\citep{kim2018disentangling}
  or GAN~\citep{chen2016infogan}.
  Yet in the irregular domain, 
  unfortunately, state-of-the-art graph generative models 
  are only applicable for generating small graphs or
  larger ones without features.
  {Moreover, these models comprise a sequential generation step,
  making it infeasible to be integrated into the graph disentangling frameworks.}
  To this end, we propose a graph edit-distance based metric,
  which bypasses the generation step
  and estimates the similarity between the factor graphs and the ground truth.
  
  %However, this is not the case in the field of structural domain, since
  %current GCN models can only generate a small graph, or a larger graph without  features. We instead propose a new metric inspired by graph edit distance to   measure the quality of disentanglement of graph data.
\end{itemize}
  
We conducted experiments on five datasets
in various domains,
and demonstrate that the proposed FactorGCN
yields state-of-the-art performances 
for both disentanglement and
downstream tasks.
This indicates that,
even putting side its disentangling capability, 
FactorGCN may well serve as a general GCN framework. 
Specifically, on the ZINC dataset~\citep{jin2018junctionZINC},
FactorGCN outperforms other methods by a large margin,
and, without {the bond information of the edges},
FactorGCN achieves a performance on par with the state-of-the-art
method that explicitly {utilizes}
edge-type information.

\iffalse
For two of them, ground truhts of the 
disentangled factor graphs are available;
on these two datasets,
FactorGCN performs consistently the best
in terms of both the disentanglement performance 
and the downstream task performance.
The other three datasets are from 

social network 
and bioinformatics graph, 
on which FactorGCN 
achieves the state-of-the-art performance,
showing that it is ready 
to be used as a general GCN framework.  
Specifically, on the ZINC dataset, our method
outperforms the other methods by a large margin
and achieve a similar performance as the state-of-the-art
method that explicitly \emph{utilizes} the type information of edges,
indicating that the disentangled factor graphs can 
indeed boost results of the downstream tasks.
\fi

\section{Related Work}

\textbf{Disentangled representation learning}. 
Learning disentangled representations has recently
emerged as a significant task towards 
interpretable AI~\citep{yang2020ECCV,Song_2020_CVPR}.
Unlike earlier attempts that rely on 
handcrafted disentangled representations
or variables~\citep{WangECCV14,WangTPAMI16}, 
most of the recent works
in disentangled representation learning are based on the architecture
of auto-encoder~\citep{higgins2017betaVAE,feng2018dual,bouchacourt2018multi,burgess2018understanding,wang2017tag,kim2018disentangling} 
or generative
model~\citep{chen2016infogan,zhao2017learning,siddharth2017learning}.
One mainstream auto-encoder approach is to constrain
the latent feature generated from the encoder to make it independent
in each dimension. For example, VAE~\citep{kingma2013autovae}
constrains the distribution of the latent features to Gaussian;
$\beta$-VAE\citep{higgins2017betaVAE}
enlarges the weight of the KL divergence term to 
%make the distribution
%of the latent feature is closer to the Gaussian one;
balance the independence constraints and reconstruction accuracy;
\citep{schmidhuber1992learning} disentangles the latent features by
ensuring that each block of latent features cannot be predicted
from the rest; 
DSD~\citep{feng2018dual} swaps some of the latent features
twice to achieve semi-supervised disentanglement. 
For the generative model, extra information is introduced during the
generation. For example, InfoGAN~\citep{chen2016infogan} adds the class code to 
the model and maximizes the mutual information between the
generated data and the class code.

\textbf{Graph convolutional network}.
Graph convolutional network~(GCN) has shown its potential in the
non-grid domain~\citep{xu2018powerful,Qiu2020ECCV,li2018combinatorial,yang2020distilling,monti2017geometricMoNet,ijcai_spagan}, achieving promising results on various type of 
structural data, like citation graph~\citep{velickovic2018graphgat}, 
social graph~\citep{kipf2017semi}, 
and relational graph~\citep{schlichtkrull2018modeling}.
Besides designing GCN to better extract information from
non-grid data, there are also a couple of works that explore the 
disentangled GCNs~\citep{ma2019learning,liu2019independence}. 
DisenGCN~\citep{ma2019disentangled} adopts 
neighbour routine to divide the neighbours of the node 
into several mutually exclusive parts.
IPGDN~\citep{liu2019independence} improves DisenGCN
by making the different parts of the embedded feature
independent. Despite results of the previous works, there
remain still several problems: 
the disentanglement is 
in the node level, which does not consider the information of
the whole graph,
and there is no quantitative metrics to evaluate
the performance of disentanglement.

\section{Method}

%The basic component of the proposed FactorGCN is the disentangle layer,
%as shown in Fig.~\ref{fig:factorlayer}.

In this section, we will give a detailed description 
about the architecture of FactorGCN, whose basic component 
is the disentangle layer, as shown in Fig.~\ref{fig:factorlayer}.

%the architecture of the disentangle layers
%\xw{and the architecture of the FactorGCN.}

\subsection{{Disentangling Step}}
The goal of this step is to factorize the input graph into several factor graphs.
To this end, we treat the edges equally across the whole graph.
The mechanism we adopt to
generate these factorized coefficient is
similar to that of graph attention network~\citep{velickovic2018graphgat}.
We denote the input of the disentangle layer as
$\mathbf{h} = \{h_0, h_1, ..., h_n\}, h_i \in \mathcal{R}^F$ 
and 
$\mathbf{e} = \{e_0, e_1, ..., e_m\}, e_k = (h_i, h_j)$.
$\mathbf{h}$ denotes the set of nodes with feature of $F$ dimension, and 
$\mathbf{e}$ denotes the set of edges. 
%indicating whether two nodes are connected. 

The input nodes are transformed to a new space,  done by multiplying the 
features of nodes with a linear transformation matrix 
$\mathbf{W} \in \mathcal{R}^{F^\prime \times F}$. 
This is a standard operation in most GCN models, which increases the capacity of the model.
The transformed features are then used to 
generate the factor coefficients as follows
\begin{equation}
    E_{ije} = 1 / \left(1 + e^{-\Psi_e (h^\prime_{i}, h^\prime_{j}) } \right); h^\prime=\mathbf{W} h,
    \label{eq:1}
\end{equation}
where $\Psi_{e}$ is the function that takes the features of
node $i$ and node $j$ as input and computes the attention score of the edge
for factor graph $e$, 
and takes the form of  an one-layer MLP
in our implementation; 
$E_{ije}$ then can be obtained by normalizing the attention score 
to $[0, 1]$, representing the coefficient of edge from node $i$ to node $j$
in the factor graph $e$;
{$h^\prime$ is the transformed node feature, shared
across all functions $\Psi_{*}$.} Different from most previous 
{forms of attention-based GCNs} that normalize
the attention coefficients among all the neighbours of nodes, 
our proposed model generates these coefficients directly
{as the factor graph}.

Once all the coefficients are computed,
a factor graph $e$ can be represented by its own $E_e$,
which will be used for the next aggregation step. 
However, without any other
constrain, some of the generated factor graphs may contain a similar
structure, degrading the disentanglement performance and 
capacity of the model. We therefore introduce an additional
head in the disentangle layer, aiming to avoid the 
degradation of the generated factor graphs.

The motivation of the additional head is that, a well
disentangled factor graph should have enough information to 
be distinguished from the rest, only based on its
structure. 
{Obtaining the solution that all the disentangled
factor graphs differ from each other to the
maximal degree, unfortunately,  is not trivial.}
We thus approximate the solution by 
giving unique labels to the factor graphs
and optimizing the factor graphs
as a graph classification problem. 
%{To this end, we give unique labels to the factor graphs} The virtual labels are created from the \xw{orders} %of the factor graphs, 
%meaning that the number of classes 
%is taken to be the number of factor graphs,
%and the class of each factor graph will be the same as its \xw{order}.
Our additional head will serve as a {discriminator, shown in Eq.~\ref{eq:2}}, to distinguish which label a given graph has:
\begin{small}
\begin{equation}
    G_e = {\rm Softmax}\Big( f \big({\rm Readout}(\mathcal{A}(\mathbf{E}_{e}, \mathbf{h^\prime}) ) \big) \Big).
    \label{eq:2}
\end{equation}
\end{small}
 
The discriminator contains 
a three-layer graph auto-encoder $\mathcal{A}$, which takes the transformed feature
$\mathbf{h^\prime}$ and the generated attention coefficients of factor graph $\mathbf{E}_e$
as inputs, and generates the new node features. 
These features are then readout to generate
the representation of the whole factor graph. 
Next, the feature vectors will be sent to
a classifier with one fully connected layer.
Note that all the factor graphs share the 
same {node features}, making sure that the 
information discovered by the discriminator only
comes from the difference among the structure of
the factor graphs.
More details about the discriminator architecture
can be found in the supplementary materials.

The loss used to train the discriminator 
is taken as follows:
\begin{small}
\begin{equation}
    \mathcal{L}_{d} = - \frac{1}{N} \sum_i^N \left( \sum_{c=1}^{N_e} \mathbbm{1}_{e=c} log(G_i^e[c])  \right),
    \label{eq:disloss}
\end{equation}
\end{small}\noindent
where $N$ is the number of training samples, 
set to be the number of input graphs 
multiplies by the number of factor graphs; 
$N_e$ is the number of factor
graphs; $G_i^e$ is the distribution
of sample $i$ and $G_i^e[c]$ represents the
probability that the generated factor graph has label $c$. 
$\mathbbm{1}_{e=c}$ is an indicator function, taken to be one
when the predicted label is correct.

\subsection{{Aggregation Step}}
%\xw{In the disentangling step, the input graph is divided into 
%several factor graphs and optimized to be 
%as diverge as possible.}
As the factor graphs derived from the disentangling step
is optimized to be as diverse as possible,
in the aggregation step, we will use the generated factor graphs
to aggregate information in different structural spaces.

This step is similar as the most GCN models, where
the new node feature is generated by taking the weighted sum of its
neighbors. Our aggregation mechanism is based on
the simplest one, which is used in GCN~\citep{kipf2017semi}. 
The only difference is that the aggregation will take place independently
for each of the factor graphs. 

The aggregation process is formulated as
\begin{small}
\begin{equation}
    h^{(l+1)_e}_i = \sigma(\sum_{j\in \mathcal{N}_i} E_{ije} / c_{ij} h^{(l)}_j \mathbf{W}^{(l)} ), 
    c_{ij} = \left( |\mathcal{N}_i||\mathcal{N}_j| \right)^{1/2},
    \label{eq:agg}
\end{equation}
\end{small}\noindent
where $h^{(l+1)_e}_i$ represents the new feature for node
$i$ in $l+1$ layer aggregated 
{from} the factor graph $e$; $\mathcal{N}_i$ represents all
the neighbours of node $i$ in the input graph;
$E_{ije}$ is the coefficient of the edge from node $i$ to
node $j$ in the factor graph $e$; $c_{ij}$ is the
normalization term that is computed according to
the degree of node $i$ and node $j$; 
$\mathbf{W}^{(l)}$ is a linear transformation matrix, 
which is the same as the matrix used in the disentangling step.

Note that although we use all the neighbours of a node
in the input graph to aggregate information, 
{some of them are making no contribution if the corresponding
coefficient in the factor graph is zero.}

\subsection{{Merging Step}}
Once the aggregation step is complete,
different factor graphs will lead to 
different features of nodes. 
We merge these  features generated from 
different factor graphs by applying
\begin{small}
\begin{equation}
    h^{(l+1)}_i = ||^{N_e}_{e=1} h^{(l+1)_e}_i,
    \label{eq:merge}
\end{equation}
\end{small}\noindent
where $h^{(l+1)}_i$ is the output feature of node $i$; $N_e$
is the number of factor graphs; $||$ represents the 
concatenation operation.

\subsection{Architecture}

We discuss above the design of one disentangle layer, 
which contains three steps. The FactorGCN model 
we used in the experimental section contains
several such disentangle layers, increasing
the power of expression. Moreover, by setting different
number of factor graphs in different layers, 
the proposed model can disentangle the input data
in a hierarchical manner.

The total loss to train FactorGCN model is $\mathcal{L} = \mathcal{L}_{t} + \lambda * \mathcal{L}_{d} $. $\mathcal{L}_{t}$ is the loss of
the original task, which is taken to be 
a binary cross entropy loss for multi-label
classification task, cross entropy loss for
multi-class classification task, or L1 loss for regression task.
$\mathcal{L}_{d}$ is the loss 
of the discriminator we mentioned above. $\lambda$ is the 
weight to balance these two losses.

\section{Experiments}

In this section, 
we show the effectiveness of the proposed FactorGCN,
and provide discussions
on its various components as well as 
the sensitivity with respect to the
key hyper-parameters.
More results can be found in the supplementary materials. 

\subsection{Experimental setups}

{\textbf{Datasets}. }
Here, we use six datasets to evaluate the 
effectiveness of the proposed method.
The first one is a synthetic dataset 
that contains a fixed number of predefined graphs
as factor graphs. The second one is the ZINC dataset~\citep{dwivedi2020benchmarkinggnn} 
built from molecular graphs.
The third one is Pattern dataset~\citep{dwivedi2020benchmarkinggnn},
which is a large scale dataset for node classification task.
The other three are widely used graph classification datasets
include social networks~(COLLAB,IMDB-B) 
and bioinformatics graph~(MUTAG)~\citep{yanardag2015deepgin}.
To generate the synthetic dataset that contains $N_e$ factor graphs,
we first generate $N_e$ predefined graphs, 
which are the well-known graphs like Tur\'an graph, house-x graph, 
and balanced-tree graph. 
We then choose half of them and pad them with isolated nodes to 
make the number of nodes to be 15.
The padded graphs will be merged together as a training sample.
The label of the synthetic data is a binary vector, with the dimension $N_e$.
Half of the labels will be set to one according to
the types of graphs that the sample generated from, and
the rest are set to zero.
More information about the datasets can be found
in the supplemental materials.
% Here, we use \xwc{five datasets} to evaluated the effectiveness of the proposed method. One is synthetic dataset and the other one is real-world dataset. 
% The synthetic dataset is build to contains several factor graphs, which can
% be control to evaluation the proposed method thoroughly.
% To generate the synthetic dataset that contains N factor graphs,
% we first generate $N$ predefined graphs, these predefined graphs
% are from the well-known graphs like Tur\'an graph, house-x graph, 
% and balanced-tree graph. Then, we choose half of them and pad them
% with isolated nodes to make the number of nodes equal to 15.
% The padded graph when union together to form a training sample.
% The label of the synthetic data is a binary vector, with the dimension $N$.
% Half of the label will be set to one according to
% the types of graphs where the sample generated from, the rest is set to zero.
% We generate 20,000 samples, among them, 14,000 for training, 
% 2,000 for validation, 4,000 for testing. Micro-F1 is used to evaluation the
% performance.

% We adopt ZINC~\citep{dwivedi2020benchmarkinggnn} as the real-world dataset,
% which is build from the molecular graphs. The target of the dataset
% is to regress a molecular property~\citep{jin2018junctionZINC}. The node of
% the graph represents atom and edge represents bond. We use 10,000 for traininn,
% 1,000 for validation and 1,000 for testing. Mean absolute error~(MAE) is used
% to evaluate the performance. 
% We also adopt three more graph datasets to evaluate
% the performance of FactorGCN as a general GCN framework.

\textbf{Baselines}. We adopt several methods, 
including state-of-the-art ones, as the baselines. 
Among all, 
MLP is the simplest one, which contains multiple fully connected layers.
Although this method is simple, it can in fact perform well when comparing
with other methods that consider the structural information.
We use MLP to check whether the other compared methods  benefit
from using the structural information as well.
GCN aggregates the information in the graph according to
the laplacian matrix of the graph,
which can be seen as a fixed weighted sum on  
the neighbours of a node. 
GAT~\citep{velickovic2018graphgat} extends the idea of GCN by 
introducing the attention mechanism.
The weights when doing the aggregation is computed dynamically according to
all the neighbours. For the ZINC dataset, we also add MoNet~\citep{monti2017geometricMoNet}
and GatedGCN$_E$~\citep{dwivedi2020benchmarkinggnn}
as  baselines. The former one is the state-of-the-art method that does not
use the type information of edges while the latter one is the state-of-the-art
one that uses additional edge information.
Random method is also added to provide the result of random guess for reference. For the other three graph datasets, we add
non DL-based methods~(WL subtree, PATCHYSAN, AWL) and 
DL-based methods~(GCN, GraphSage~\citep{hamilton2017inductive}, GIN)
as baselines. DisenGCN~\citep{ma2019disentangled} and 
IPDGN~\citep{liu2019independence}
are also added.

\textbf{Hyper-parameters}.
For the synthetic dataset, Adam optimizer is used with a
learning rate of 0.005, 
the number of training epochs is set to 80, 
the weight decay is set to 5e-5.
The row of the adjacent matrix of the generated synthetic 
graph is used as the feature of nodes. 
The negative slope of LeakyReLU for GAT model is set to 0.2, 
which is the same as the original setting. 
The number of hidden layers for all models is set to two. 
The dimension of the hidden feature is set to 32 when 
the number of factor graphs is no more than four
and 64 otherwise. 
The weight for the loss of discriminator in FactorGCN is set to 0.5.

For the molecular dataset, the dimension of the hidden feature is set to 144 
for all methods and the number of layers is set to four. 
Adam optimizer is used with a learning rate of 0.002. 
No weight decay is used. 
$\lambda$ of FactorGCN is set to 0.2. All the methods
are trained for 500 epochs. The test results are obtained using the model with 
the best performance on validation set. For the other three datasets,
three layers FactorGCN is used.

\begin{figure}
  \centering
  \includegraphics[width=0.95\linewidth]{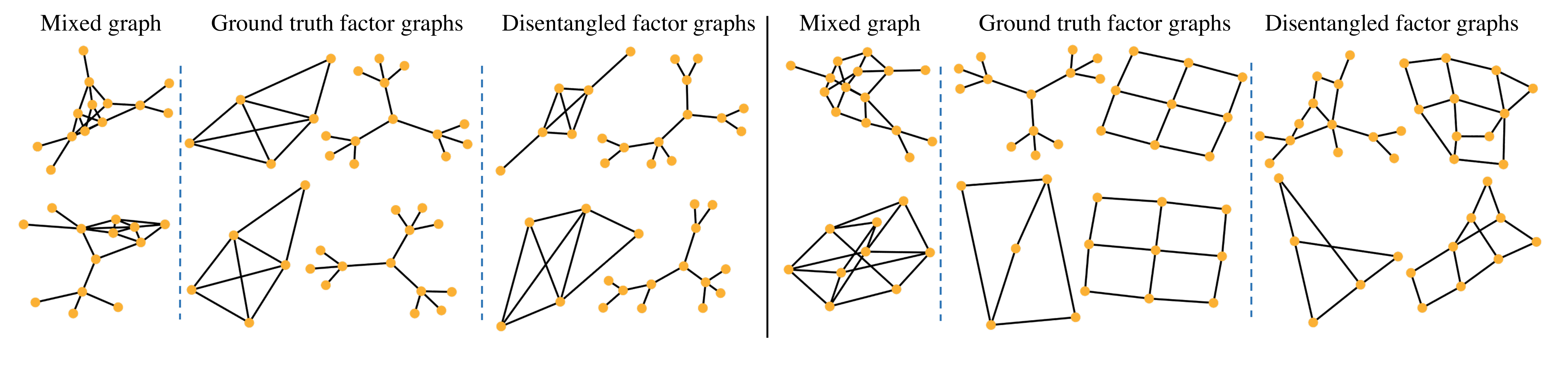}
   \vspace{-1.3em}
  \caption{Examples of the disentangled factor graphs on the synthetic dataset. 
  The isolated nodes are eliminated for a better visualization.}
  \label{fig:synthetic_vis}
  \vspace{-1.2em}
\end{figure}

\subsection{Qualitative Evaluation}
We first provide the qualitative evaluations of disentanglement performance,
including the visualization of the disentangled factor graphs and 
the correlation analysis of the latent features.

\textbf{Visualization of disentangled factor graphs}.
To give an intuitive understanding of the 
disentanglement. We provide in Fig.~\ref{fig:synthetic_vis}
some examples of the generated factor graphs. 
We remove the isolated nodes and visualize
the best-matched factor graphs with  ground truths.
More results and analyses can be found in the supplemental
materials.

\begin{figure}[ht]
  \centering
  \includegraphics[width=0.93\linewidth]{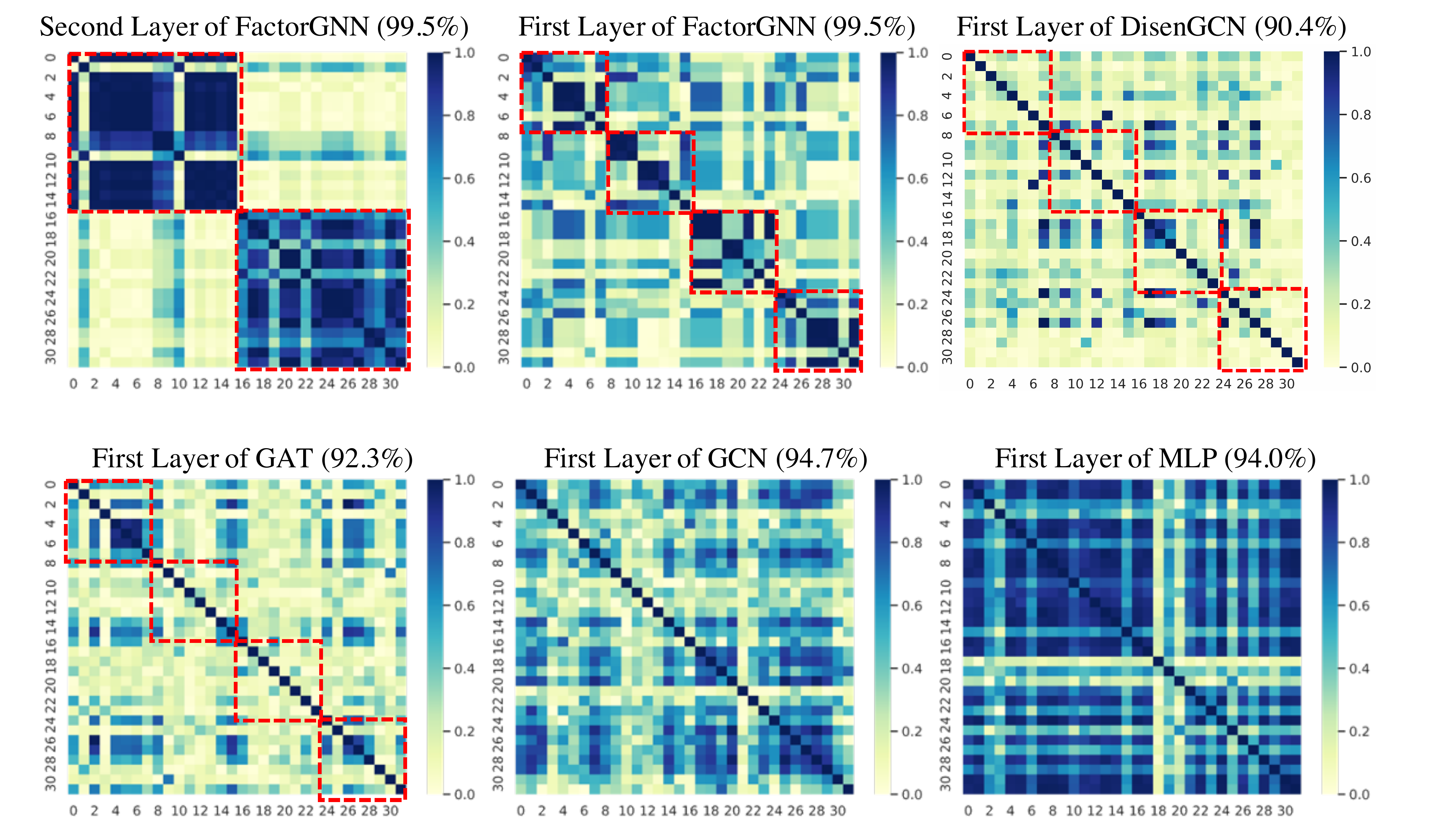}
%   \vspace{-0.4em}
  \caption{Feature correlation analysis. The hidden features are obtained from
  the test split using the pre-trained models on the synthetic dataset.
  It can be seen that the features generated 
  from FactorGCN present a more block-wise
  correlation pattern, indicating that 
  the latent features have indeed been disentangled.
  We also show the classification performance in brackets.}
    \label{fig:synthetic_corr}
%   \vspace{-0.2em}
\end{figure}

\textbf{Correlation of disentangled features}.
Fig.~\ref{fig:synthetic_corr} shows the correlation 
analysis of the latent features
obtained from several pre-trained models on the synthetic dataset.  
It can be seen that also GCN and MLP models can achieve 
a high performance in the downstream task, and
their latent features are hidden entangled. 
GAT gives {more} independent latent features 
but the performance is degraded in the 
original task. FactorGCN is able to extract the highly independent
latent features and meanwhile achieve a better performance in the downstream task.

\subsection{Quantitative Evaluation}
The quantitative evaluation focuses on two parts,
the performance of the downstream tasks 
and that of the disentanglement.

\textbf{Evaluation protocol}. 
For the downstream tasks, 
we adopt the corresponding metrics to evaluate,
i.e., Micro-F1 for the multi-label classification task,
mean absolute error~(MAE) for the regression task.
We design two new metrics to evaluate the disentanglement
performance on the graph data.
The first one is graph edit distance on edge~(GED$_{E}$).
This metric is inspired by the traditional graph edit distance~(GED).
Since the input graph already provides the information
about the order of nodes, the disentanglement of 
the input data, in reality, 
only involves the changing of edges.
Therefore, we restrict the GED by only 
allowing adding and removing the edges, and 
thus obtain a score of 
GED$_{E}$ by Hungarian match 
between the generated factor graphs
and the ground truth.

Specifically, for each pair of the 
generated factor graph and the
ground truth graph, we first convert the continuous value in the 
factor graph to 1/0 value by setting the threshold to make
the number of edges in these two graphs are the same.
Then, GED$_{E}$s can be computed for every such combination.
Finally, Hungarian match is adopted to obtain the best bipartite matching results as the GED$_{E}$ score. 

Besides the GED$_{E}$ score, we also
care about the consistency of the generated factor graph. 
In other words, the best-matched pairs between the generated factor graphs and the ground truths, optimally, should be 
identical across all samples.
We therefore introduce the second metric named as consistency score~(C-Score), related to GED$_{E}$. 
C-Score is computed as the 
average percentage of the 
most frequently matched factor graphs.
The C-score will be one 
if the ground truth graphs are always matched
to the fixed factor graphs. 
A more detailed description of evaluation
protocol can be found in the supplemental materials.

\begin{table}
  \caption{Performance on synthetic dataset. The four methods are  evaluated in terms of 
  the classification and the disentanglement performance. Classification performance
  is evaluated by Micro-F1 and disentanglement performance is measured by GED$_E$ and C-Score. For each method,
  we run the experiments five times and report the mean and std. Random method generates four factor graphs. 
  GAT\_W/Dis represents GAT model with
  the additional discriminator proposed in this paper.}
  \label{tab:synthetic}
  \centering
  \scalebox{0.71}{
  \begin{tabular}{lccccccc}
    \toprule
     & MLP & GCN & GAT & GAT\_W/Dis & DisenGCN & FactorGCN~(Ours) & Random\\
    \midrule
     Micro-F1 $\uparrow$  & 0.940 $\pm$ 0.002 & 0.947 $\pm$ 0.003 & 0.923 $\pm$ 0.009  &  0.928 $\pm$ 0.009  & 0.904$\pm$0.007 & \textbf{0.995 $\pm$ 0.004} & 0.250 $\pm$ 0.002 \\
    GED$_{E}$ $\downarrow$  & - & - & 12.59 $\pm$ 3.00 &  12.35 $\pm$ 3.86
    & \textbf{10.54$\pm$4.35} & \textbf{10.59 $\pm$ 4.37} & 32.09 $\pm$ 4.85 \\
    C-Score $\uparrow$  & - & - & 0.288 $\pm$ 0.064  & 0.274 $\pm$ 0.065
    & 0.367$\pm$0.026 & \textbf{0.532 $\pm$ 0.044} & 0.315 $\pm$ 0.002 \\
    \bottomrule
  \end{tabular}
  }
%   \vspace{-0.2em}
\end{table}

\textbf{Evaluation on the synthetic dataset}. We first evaluate the disentanglement 
performance on a synthetic dataset. The results are shown in Tab.~\ref{tab:synthetic}.
Although MLP and GCN achieve  good classification performances, they are not capable of disentanglement. 
GAT  disentangles the input by using multi-head attention,
but the performance of the original task is degraded. 
Our proposed method,
on the other hand, achieves a much better performance in terms of both disentanglement and
the original task.
We also evaluate the compared methods on the synthetic dataset with various numbers of 
factor graphs, shown in Tab.~\ref{tab:synthetic_various}. 
As the number of
latent factor graphs increase, the performance gain of the FactorGCN becomes large.
However, when the number of factor graphs becomes too large, 
the task will be more challenging,
yielding lower performance gains.

\begin{table}
  \caption{Classification performance on synthetic graphs with different numbers of factor graphs.
  We change the total number of factor graphs and generate five synthetic datasets.
  When the number of factor graphs increases, the performance gain of FactorGCN becomes larger.
  However, as the number of factor graphs becomes too large, disentanglement
  will be more challenging, yielding {lower performance gains}.}
  \label{tab:synthetic_various}
  \centering
  \scalebox{0.92}{
  \begin{tabular}{cccccc}
    \toprule
     \multirow{2}{*}{Method} & \multicolumn{5}{c}{Number of factor graphs} \\
     \cmidrule(r){2-6}
      & 2 & 3 & 4 & 5  & 6 \\
      \midrule
      MLP       & 1.000 $\pm$ 0.000 & 0.985 $\pm$ 0.002 & 0.940 $\pm$ 0.002  & 0.866 $\pm$ 0.001 & 0.809 $\pm$ 0.002 \\
      GCN       & 1.000 $\pm$ 0.000 & 0.984 $\pm$ 0.000 & 0.947 $\pm$ 0.003 & 0.844 $\pm$ 0.002 & 0.765 $\pm$ 0.001 \\
      GAT       & 1.000 $\pm$ 0.000 & 0.975 $\pm$ 0.002 & 0.923 $\pm$ 0.009 & 0.845 $\pm$ 0.006 & 0.791 $\pm$ 0.006 \\
      FactorGCN & 1.000 $\pm$ 0.000 & \textbf{1.000 $\pm$ 0.000} & \textbf{0.995 $\pm$ 0.004} & \textbf{0.893 $\pm$ 0.021} & \textbf{0.813 $\pm$ 0.049} \\
    \bottomrule
  \end{tabular}
  }
\end{table}

\begin{table}
  \caption{Performance on the ZINC dataset. FactorGCN outperforms the compared methods 
  by a large margin, with the 
  capability of disentanglement. 
  Note that our proposed method even achieves a similar performance as
  GatedGCN$_E$, the state-of-the-art method on ZINC dataset that explicitly uses additional
  edge information. }
  \label{tab:zinc}
  \centering
  \scalebox{0.70}{
  \begin{tabular}{c|cccccc|c}
    \toprule
     & MLP & GCN & GAT & MoNet & DisenGCN & FactorGCN~(Ours) & GatedGCN$_E$\\
    \midrule
    MAE $\downarrow$ & 0.667 $\pm$ 0.002 & 0.503 $\pm$ 0.005 & 0.479 $\pm$ 0.010 & 0.407 $\pm$ 0.007 
    & 0.538$\pm$0.005  & \textbf{0.366 $\pm$ 0.014}  & \textit{0.363 $\pm$ 0.009}\\
    GED$_{E}$ $\downarrow$   & -  & - & 15.46 $\pm$ 6.06 & - 
    & 14.14$\pm$6.19 & \textbf{12.72 $\pm$ 5.34} &- \\
    C-Score $\uparrow$  & -  & - & 0.309 $\pm$ 0.013 & - 
    & 0.342$\pm$0.034 & \textbf{0.441 $\pm$ 0.012}  &- \\
    \bottomrule
  \end{tabular}
  }
\end{table}

\begin{table}
  \caption{Accuracy~(\%) on three graph classification datasets. 
  FactorGCN performances on par with or better 
  than the state-of-the-art
  GCN models. We highlight the best DL-based methods and non DL-based methods
  separately. FactorGCN uses the same hyper-parameters for all datasets.}
  \label{tab:other}
  \centering
  \scalebox{0.85}{
  \begin{tabular}{cccc|cccc}
    \toprule
      & WL subtree & PATCHYSAN & AWL & GCN & GraphSage & GIN & FactorGCN \\
    \midrule
    IMDB-B  & 73.8 $\pm$ 3.9 & 71.0 $\pm$ 2.2 & \textbf{74.5 $\pm$ 5.9}& 74.0 $\pm$ 3.4 &72.3 $\pm$ 5.3 & \textbf{75.1 $\pm$ 5.1} & \textbf{75.3 $\pm$ 2.7} \\
    COLLAB & \textbf{78.9 $\pm$ 1.9} & 72.6 $\pm$ 2.2 & 73.9 $\pm$ 1.9 & 79.0 $\pm$ 1.8 & 63.9 $\pm$ 7.7 & 80.2 $\pm$ 1.9 & \textbf{81.2 $\pm$ 1.4} \\
    MUTAG & 90.4 $\pm$ 5.7 & \textbf{92.6 $\pm$ 4.2} & 87.9 $\pm$ 9.8 & 85.6 $\pm$ 5.8 & 77.7 $\pm$ 1.5 & \textbf{89.4 $\pm$ 5.6} & \textbf{89.9 $\pm$ 6.5}\\
    \bottomrule
  \end{tabular}
  }
\end{table}

\begin{table}[ht]
  \caption{Accuracy~(\%) on the Pattern dataset 
  for node-classification task. 
  FactorGCN achieves the best performance, 
  showing its ability to serve as a general GCN framework.}
  % This dataset contains 14K graphs with a total 1,664,491 nodes.
  \label{tab:pattern}
  \centering
  \scalebox{0.86}{
  \begin{tabular}{ccccccc}
    \toprule
    GCN & GatedGCN & GIN & MoNet & DisenGCN & IPDGN & FactorGCN \\
    \midrule
    63.88 $\pm$ 0.07 & 84.48 $\pm$ 0.12 & 85.59 $\pm$ 0.01 
    & 85.48 $\pm$ 0.04 & 75.01 $\pm$ 0.15 & 78.70 $\pm$ 0.11 & \textbf{86.57 $\pm$ 0.02} \\
    \bottomrule
  \end{tabular}
  }
\end{table}

\begin{figure}[ht]
  \centering
  \includegraphics[width=0.99\linewidth]{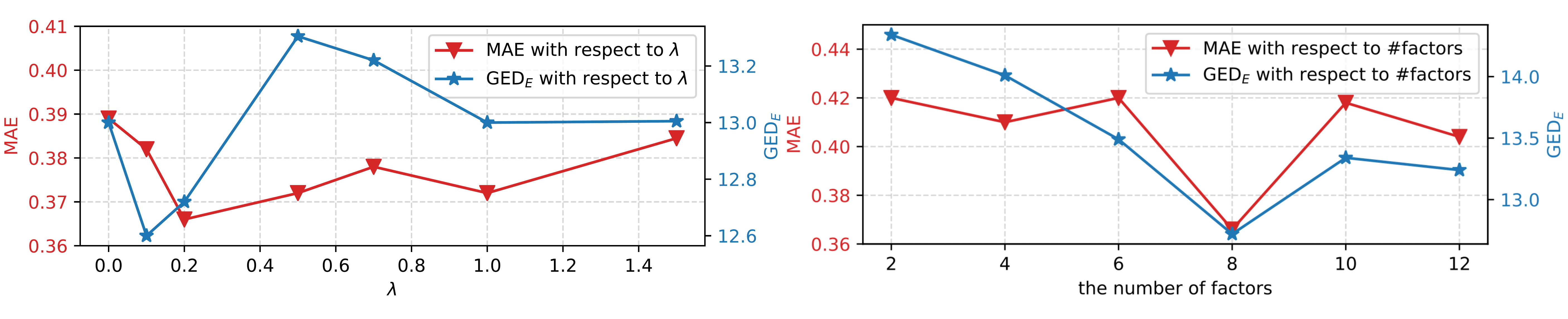}
  %\vspace{-1em}
  \caption{The influence of the balanced weight $\lambda$ and 
  the number of factor graphs.}
  %\vspace{-1.5em}
    \label{fig:sens}
\end{figure}

\textbf{Evaluation on the ZINC dataset}. 
%We adopt the ZINC dataset as a real-world data to evaluate all the compared methods. 
For this dataset, the type information of edges is 
hidden during the training process,
and is serve as the ground truth to evaluate the performance of
disentanglement. Tab.~\ref{tab:zinc} shows the results. The proposed method
achieves the best performance on both the disentanglement
and the downstream task. 
We also show the state-of-the-art method GatedGCN$_E$
on this dataset on the right side of Tab.~\ref{tab:zinc}, which utilizes
the type information of edges during the training process.
Our proposed method, without any additional edge information,
achieves truly promising results that are
to that of GatedGCN$_E$, which {needs the bond information
of edges during training.}

\textbf{Evaluation on more datasets}. 
To provide a thorough understanding of the proposed method, 
We also carry out evaluations on three widely 
used graph classification datasets and one node classification dataset
to see the performances of FactorGCN as a general GCN framework.
The same 10-fold evaluation protocol
as~\citep{xu2018powerful} is adopted. 
Since there are no ground truth factor graphs, 
we only report the accuracy, shown in 
Tab.~\ref{tab:other} and Tab.~\ref{tab:pattern}. 
Our method achieves consistently the best performance,
showing the potential of the FactorGCN 
as a general GCN framework, even putting aside its
disentangling capability.
More details about the evaluation protocol, 
the setup of our method, and the statistic information about these datasets 
can be found in the supplemental materials.

\subsection{Ablation and sensitivity analysis}

We show in Fig.~\ref{fig:sens} the ablation study and sensitivity analysis
of the proposed method.
When varying $\lambda$, the number of factors 
is set to be eight; 
when varying the number of factors , 
$\lambda$ is set to be 0.2.
As can be seen from the left figure,
the performance of both the disentanglement and the 
downstream task will degrade without the discriminator.
The right figure shows the relations between the performance
and the number of factor graphs we used in FactorGCN.
Setting the number of factor graphs 
to be slightly larger than that of the ground truth,
in practice, leads to a better performance.

\section{Conclusion}
We propose a novel GCN framework, termed as FactorGCN,
which achieves graph convolution through 
graph-level disentangling. Given an input graph, FactorGCN
decomposes it into several interpretable factor graphs, each of which denotes an underlying interconnections
between entities, and then carries out topology-aware convolutions on each such factor graph to produce the final node features. The node features, derived under the explicit disentangling,  are therefore block-wise explainable
and beneficial to the downstream tasks. 
Specifically, FactorGCN enables multi-relation 
disentangling, allowing information propagation
between two nodes to take places 
in disjoint spaces.
We also introduce two new metrics to 
measure the graph disentanglement
performance quantitatively. 
FactorGCN outperforms other methods
on both the disentanglement and the downstream tasks, indicating
the proposed method is ready to serve as a general GCN framework
with the capability of graph-level disentanglement.

\section*{Acknowledgments}
This work is supported by the startup funding of 
Stevens Institute of Technology.

\section*{Broader Impact}
In this work we introduce a GCN framework,
termed as FactorGCN,
that explicitly accounts for disentanglement
FactorGCN is applicable to various scenarios, 
both technical and social. 
For conventional graph-related
tasks, like node classification of the social network
and graph classification of the molecular graph, 
our proposed method can serve as a general GCN
framework.
For disentangling tasks, our method
generates factor graphs that reveal
the latent relations among entities,
and facilitate the 
further decision making process like
recommendation.
Furthermore, given sufficient data,
FactorGCN can be used as a tool to 
analyze social issues like discovering 
the reasons for the quick spread of 
the epidemic disease in some areas. 
Like all learning-based methods,
FactorGCN is not free of errors.
If the produced disentangled factor graphs 
are incorrect, for example,
the subsequent inference and  prediction results
will be downgraded,  possibly yielding
undesirable bias. 

\newpage

\small
\bibliographystyle{unsrtnat}
\bibliography{reference}

\end{document}